\documentclass{article}

\usepackage[preprint]{neurips_2025}

\usepackage[utf8]{inputenc} % allow utf-8 input
\usepackage[T1]{fontenc}    % use 8-bit T1 fonts
\usepackage{hyperref}       % hyperlinks
\usepackage{url}            % simple URL typesetting
\usepackage{booktabs}       % professional-quality tables
\usepackage{amsfonts}       % blackboard math symbols
\usepackage{nicefrac}       % compact symbols for 1/2, etc.
\usepackage{microtype}      % microtypography
\usepackage{xcolor}         % colors
\usepackage{makecell}
\usepackage{mathrsfs}
\usepackage{amsmath}
\usepackage{amssymb}        % geqslant and leqslant for better symbols
\usepackage{svg}
\usepackage{graphicx}

\usepackage{multirow} % Required for multirows

\title{FrankenBot: Brain-Morphic Modular Orchestration for Robotic Manipulation with Vision-Language Models}

\author{%
    Shiyi Wang\textsuperscript{1,}\thanks{
        These authors contribute equally.\\
    }
    ,Wenbo Li\textsuperscript{2,}$^*$\!\!\!
    ,Yiteng Chen\textsuperscript{2,}$^*$\!\!,\\
    Qingyao Wu\textsuperscript{2,\dag},
    Huiping Zhuang\textsuperscript{3,}\thanks{
        Corresponding authors: Huiping Zhuang(\href{mailto:hpzhuang@scut.edu.cn}{hpzhuang@scut.edu.cn}) and Qingyao Wu(\href{mailto:qyw@scut.edu.cn}{qyw@scut.edu.cn})
    }\\
    \textsuperscript{1}School of Future Technology, South China University of Technology \\
    \textsuperscript{2}School of Software Engineering, South China University of Technology \\
    \textsuperscript{3}Shien-Ming Wu School of Intelligent Engineering, South China University of Technology \\
}

\begin{document}

\maketitle

\begin{abstract}

Developing a general robot manipulation system capable of performing a wide range of tasks in complex, dynamic, and unstructured real-world environments has long been a challenging task. It is widely recognized that achieving human-like efficiency and robustness manipulation requires the robotic brain to integrate a comprehensive set of functions, such as task planning, policy generation, anomaly monitoring and handling, and long-term memory, achieving high-efficiency operation across all functions. Vision-Language Models (VLMs), pretrained on massive multimodal data, have acquired rich world knowledge, exhibiting exceptional scene understanding and multimodal reasoning capabilities. However, existing methods typically focus on realizing only a single function or a subset of functions within the robotic brain, without integrating them into a unified cognitive architecture. Inspired by a divide-and-conquer strategy and the architecture of the human brain, we propose FrankenBot, a VLM-driven, brain-morphic robotic manipulation framework that achieves both comprehensive functionality and high operational efficiency. Our framework includes a suite of components, decoupling a part of key functions from frequent VLM calls, striking an optimal balance between functional completeness and system efficiency. Specifically, we map task planning, policy generation, memory management, and low-level interfacing to the cortex, cerebellum, temporal lobe-hippocampus complex, and brainstem, respectively, and design efficient coordination mechanisms for the modules. We conducted comprehensive experiments in both simulation and real-world robotic environments, demonstrating that our method offers significant advantages in anomaly detection and handling, long-term memory, operational efficiency, and stability --- all without requiring any fine-tuning or retraining.
 
\end{abstract}

\begin{figure}[!ht]
    \centering
    \includegraphics[width=\linewidth]{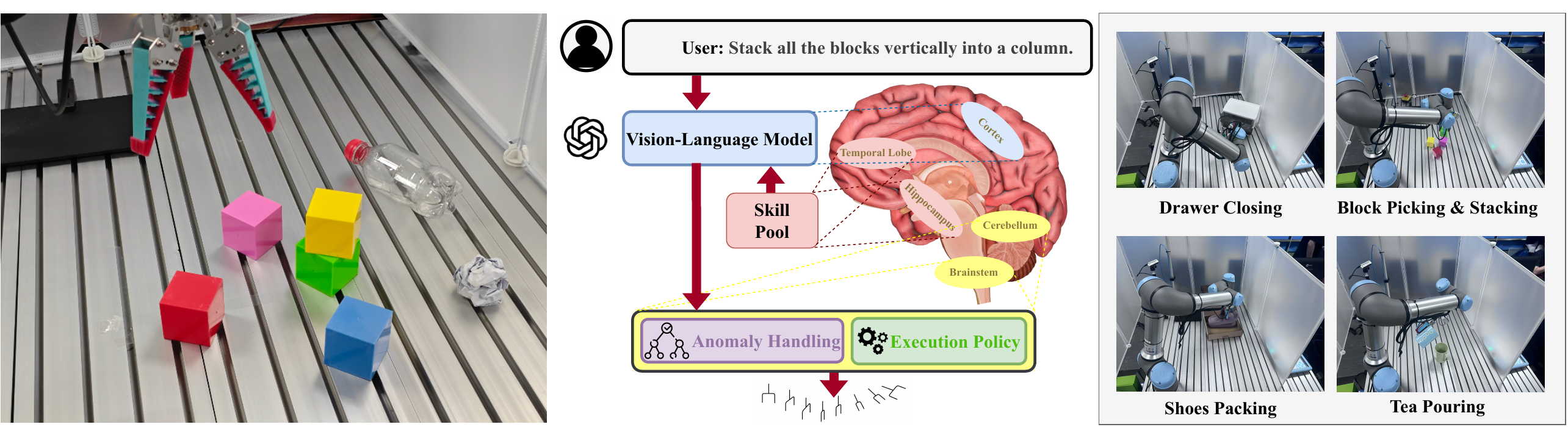}
    \caption{FrankenBot integrates task planning, policy generation, anomaly handling, and long-term memory into a unified VLM-driven cognitive architecture. In the block stacking task, FrankenBot first decomposes the task via cortex-mapped planning, then generates constrained motion policies through cerebellum-mapped optimization (blue). Concurrently, the hippocampus-mapped memory retrieves past stacking strategies (green), while brainstem-mapped reflexes (red) monitor execution stability and trigger recovery behaviors when anomalies are detected.}
    \label{fig:heading}
\end{figure}

\section{Introduction}
\label{Introduction}

Developing general robotic manipulation systems capable of performing tasks in complex, dynamic, and unstructured real-world environments has long been a challenging task. Recently, Vision-Language Models (VLMs), through large-scale pretraining, have gained rich world knowledge, demonstrating significant potential in robotic manipulation tasks. VLMs not only handle complex semantic and visual information but also enable more robust reasoning and planning across diverse scenarios, significantly reducing the reliance on high-quality action data. As a result, an increasing number of studies have explored leveraging VLMs as the core of the "robotic brain", applying them to functions such as task planning and error detection and recovery. Although VLMs excel in dialogue and static visual understanding, their pretraining primarily relies on internet text and 2D image data, and there remains a significant gap between their capabilities and the demands required for real robots to perform complex embodied tasks in 3D space. This capability gap makes it challenging for VLMs to directly adapt to dynamic environments and complex interaction requirements during embodied task deployment.

Existing VLM-driven robotic manipulation methods have already achieved the following: task planning, where VLM parses natural language instructions and generates high-level action sequences; error detection and correction, where VLM detects anomalies during task execution and replans, effectively correcting execution deviations or environmental anomalies; fine-grained action generation, where representations are first extracted, and VLM generates corresponding constraints, which are then solved to obtain the robot's action sequence. Another mainstream approach combines VLM with the Vision-Language-Action (VLA) model to build a multi-layered "robotic brain," where the upper layer provides high-level reasoning through VLM, and the lower layer handles low-level planning and execution through VLA.

While existing methods have advanced specific robotic brain functions individually, they fail to integrate these into a complete system. Current two-layer architectures demand extensive cross-modal training data and remain impractical for plug-and-play deployment. Simply combining these discrete functions would severely degrade performance, as each VLM call requires several seconds - making efficient multifunctional operation impossible. For robust real-world performance, robots need human-like brain functions (planning, error correction, and memory) operating synergistically. This work addresses the core challenge: how to build an efficient, brain-inspired framework with complete functionality that requires minimal (ideally single) VLM calls per task without additional training.

Inspired by the brain's organizational principles, we propose a VLM-driven brain-morphic robotic framework that strategically decouples core functions from frequent VLM calls. Our architecture mirrors neurobiological structures - cortical functions map to VLM-based planning and reasoning, cerebellar roles to execution control and anomaly handling, hippocampal functions to memory management, and brainstem operations to low-level hardware interfaces. Through multi-granular skill libraries, hierarchical anomaly handling, and parallel execution modules, the framework achieves comprehensive functionality while minimizing VLM interactions. This biologically-inspired design enables efficient task execution with typically just one VLM call per task, balancing system capabilities with operational performance through optimized module coordination.

Our method offers several key advantages: (1) Plug-and-play: The framework is driven by VLM, and no additional training is required for deployment. It can be quickly set up by simply connecting predefined control interfaces; (2) A comprehensive robotic brain: It incorporates essential functions such as task planning, error detection and correction, and long-term memory, thus meeting the core requirements for robotic task execution in complex environments; (3) Efficient operation: By efficiently leveraging the capabilities of VLM, the framework requires only a single VLM call during each task execution, significantly reducing computational time and resource consumption.

Our work introduces a novel VLM-driven brain-like architecture with the following contributions: First, we present a plug-and-play brain-like architecture that requires no additional training for deployment. It simply needs to implement a unified interface to call existing VLM services to drive the robot to perform complete operations. Additionally, a three-level memory mechanism has been designed, which includes a learnable multi-granular skill library that encapsulates commonly used motion primitives and composite skills in a hierarchical manner. It can adjust skill combinations based on historical task executions, improving VLM planning efficiency. Third, we introduce a multi-level anomaly handling framework, which enables a three-tier anomaly handling system from local to cloud, accurately allocating processing resources based on the complexity of the task execution’s anomalies. By combining local strategies with VLM-based judgments, this framework allows for rapid recovery and intelligent error correction. Finally, the framework enables the VLM to generate multi-threaded executable code in a single call, allowing task execution and anomaly detection to run in parallel, ensuring that in most cases, only a single VLM interaction is required for each task.

\section{Related Works}

\paragraph{Brain-like structures in Robotic Manipulation Tasks. }

A robotic system executing manipulation tasks requires a robotic brain analogous to the human brain: it must process multimodality inputs, integrate world knowledge and  reasoning capabilities, and output action sequence that cover all the functionality needed for manipulation task. The most straightforward way to build a robotic brain is an end‑to‑end, data‑driven approach: by collecting large‑scale datasets of manipulation tasks and adjusting model architectures, researchers fine‑tune pre‑trained VLMs into Vision‑Language Action (VLA) models capable of generating action sequences directly from instructions and scene observations in single inference\cite{wen2024tinyvla,kim2024openvla,liu2024rdt,brohan2022rt,brohan2023rt,gu2023rt,li2024cogact,ze20243d,ze20243d,team2024octo,cheang2024gr,wang2025roboflamingo,li2023vision,liu2025hybridvla,zhao2025cot,li2025pointvla,qu2025spatialvla}, effectively creating a fully functional robotic brain. Building on this paradigm, other works have explored more structured, brain‑like designs. Some introduce hierarchical architectures such as adopt a System 1/System 2 architecture by running parallel high‑frequency and low‑frequency inference streams\cite{bjorck2025gr00t}, while others use a two‑tier VLM/VLA design, assigning high‑level reasoning to the VLM and low‑level action generation to the VLA\cite{shi2025hi}\cite{team2025gemini}. Others layer the reasoning process temporally\cite{ji2025robobrain},  such as having the model sequentially infer task plans, predict affordances, and then produce action sequences across multiple inference steps.

\paragraph{Vision-Language Models for Robotics.}

Vision-Language Models (VLMs) have become increasingly widespread in robotic manipulation, owing to the rich scene comprehension and high‑level commonsense reasoning they acquire from large‑scale pretraining. Most prior work concentrates on leveraging pretrained VLMs for task planning and high‑level reasoning in manipulation tasks\cite{hu2023look}\cite{yang2024guiding}\cite{kumar2024open}. Another branch of research pushes VLM reasoning into more fine‑grained policy generation, beyond task planning\cite{qi2025sofar}\cite{zhao2024vlmpc}\cite{liu2025kuda}\cite{huang2023voxposer}\cite{liu2024moka}\cite{huang2024rekep}\cite{pan2025omnimanip}\cite{tang2025geomanip}. Because anomaly detection and recovery are critical for robust execution, some studies have also explored VLM‑driven monitoring and anomaly-handling strategies during task execution\cite{zhou2024code}\cite{guo2024doremi}.

In Appendix \ref{arw}, we discuss the Connections and Distinctions between our approach and existing work in these fields.

\section{Method}
\label{Method}

\begin{figure}[tb]
  \centering
  \includegraphics[width=\linewidth]{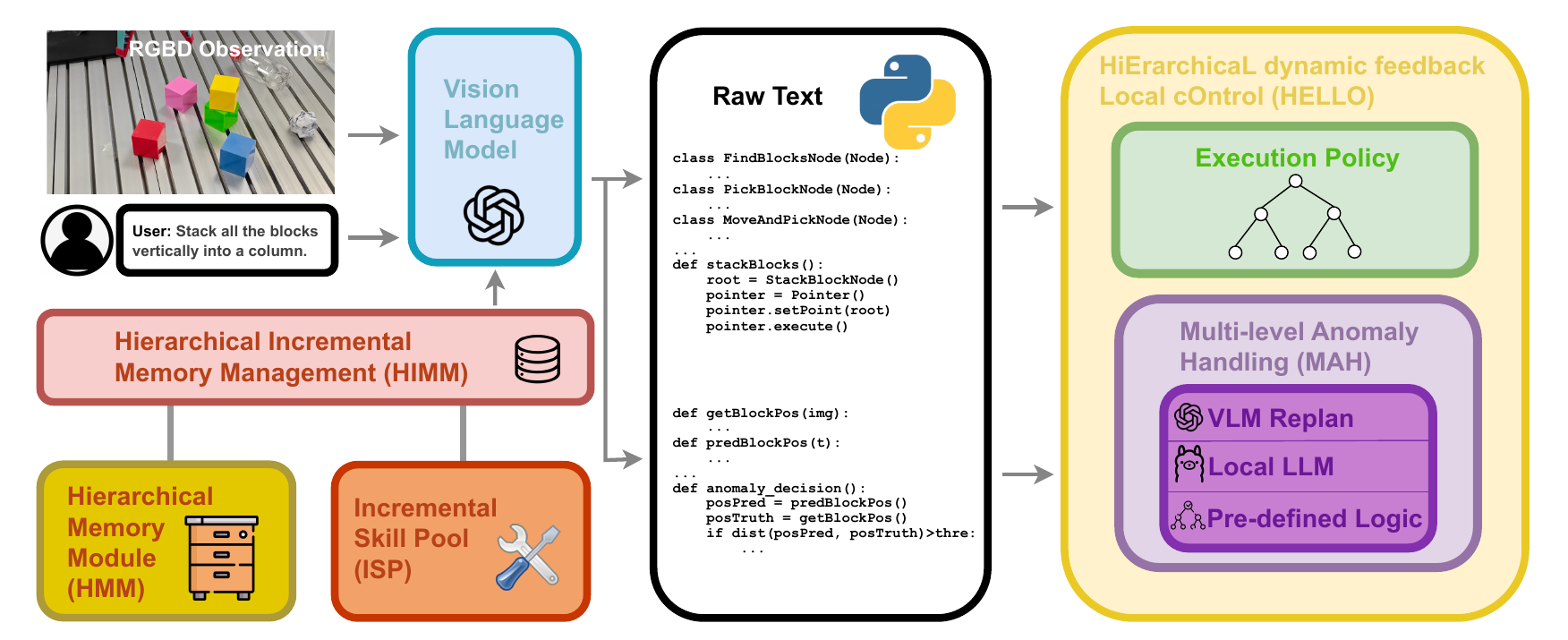}
  % \caption{Overview framework. Given natural language instruction and scene observation, VLM guides system to adaptively invoke the optimal extractor from Spatial Representation Extraction Toolkit for each task-relevant object to derive spatial representations. Low-Level Action Sequence Generator then generates robot's action sequence based on these representations and VLM-inferred constraints to complete the task.}
  \caption{Overview framework. Given real-time visual observations and instruction, the VLM dynamically generates execution policies and anomaly handling pre-defined logic that optimally selects and sequences modular skills from the Incremental Skill Pool.}
  \label{fig1main}
\end{figure}

Our framework is outlined in Fig. \ref{fig1main}, with the VLM forming the "cerebral cortex" component central to our system. In Sec. \ref{3.1}, we provide a concise formalization of the manipulation tasks and our overall architecture. Sec. \ref{3.2} details the "cerebellum" component: Hierarchical Dynamic Feedback Local Control mechanism. Sec. \ref{3.3} introduces the complementary "cerebellum" module, the Multi‑level Anomaly Handling mechanism. Sec \ref{3.4} presents the " temporal lobe-hippocampus complex" functionality  Hierarchical Incremental Memory Management which supports cross‑task information retrieval and decision enhancement.

\subsection{Problem Formulation}
\label{3.1}

In this study, we consider the task of embodied agents executing natural language instructions in real-world physical environments. Given a high-level instruction $\ell$ (e.g., "Put the apple on the table into the refrigerator") and an initial scene observation $O(0)$, the agent needs to decompose and execute it through multimodal perception and dynamic task planning. Specifically, we employ a VLM as a high-level task parser to decompose $\ell$ into an ordered set of subtasks $\{\ell_i\}_{i=1}^N$ (e.g., "Locate the apple", "Grasp the apple", "Open the fridge door", "Place the apple"). Each subtask $\ell_i$ is modeled as a task tree composed of multiple nodes $V_i=\{v_j\}_{j=1}^{M_i}$, where $M_i$ is the number of nodes of subtask $\ell_i$, with node design based on a parameterized state-transition representation.

Each node $v_j$ is defined as a 3-tuple:
$$v_j = (T_j, A_j, C_j)$$
where:
\begin{itemize}
    \item $T_j$ represents the node description text (e.g., "Grasp the apple")
    \item $A_j$ is the parameter list containing execution-specific parameters (e.g., target object coordinates, gripper force parameters)
    \item $C_j: A \rightarrow V_i$ defines the child node mapping function that maps the parameter set $A$ to subsequent node set $V_i$, enabling dynamic node selection
\end{itemize}

During execution, the agent continuously acquires environmental observations $O(t)$ through an RGB-D camera and converts them into low-dimensional scene metadata $s_t$ using predefined monitors (discussed later in Sec. \ref{3.3}). The node execution engine performs the following at each step: 1) Extract current parameters from $A_i$, 2) Execute the mapping function $C_i$, 3) Transit to the next node according to $C_i$.

To ensure robustness, an anomaly detection module $\mathcal{E}$ continuously computes state deviation $\delta_t =\mathcal{E}(s_t , u_{t})$, where $u_t$ is VLM-defined scene predictions. When $\delta_t > \tau_i$, the system either initiates VLM-predefined recovery procedures or requests the VLM to generate adjustment strategies (e.g., "If the apple is occluded, first remove the obstructing object"). This structured node modeling enables the system to both follow initial plans and adapt to environmental uncertainties through parameterized branching and anomaly handling, achieving reliable task execution in unstructured environments.

\subsection{HiErarchicaL dynamic feedback Local cOntrol(HELLO)}
\label{3.2}
% In robotics, there are multiple  approaches to task execution. A commonly adopted solution in engineering  practice is to decompose a complex task into a combination of simpler  subtasks, following the divide-and-conquer principle. Additionally,  certain operations within a task (e.g., moving above an object) are  reusable, and the complexity of expressing the entire task can be  reduced through loop and conditional branch structures. To address this,  we propose the HiErarchicaL dynamic feedback Local cOntrol(HELLO) module, which aims to achieve orderly task decomposition, dynamic  handling of different decision logics within the same task, and repeated  invocation of identical actions across different stages.

In the field of robotics, the optimization of task execution strategies typically adheres to the modular design principle. The prevalent engineering practice employs a divide-and-conquer approach, hierarchically breaking down complex tasks into combinations  of simpler subtasks. This decomposition not only reduces system  implementation complexity but also enhances reusability,  maintainability, and scalability. To systematically realize these  features, we propose the HiErarchicaL dynamic feedback Local cOntrol(HELLO) module.

% The structure of this module is illustrated in \textcolor{red}{[Figure X]}. 
It consists of a Hierarchical Execution Tree (HET) and Multi-level Anomaly Handling (MAH, discussed later in Sec. \ref{3.3}). The raw content is generated by a VLM synthesizing user instructions, initial environmental observation data $O(0)$,  and the memory module (see Sec. \ref{3.4}), then parsed by a grammar parser  into HET and MAH.

To adapt to all diverse task requirements and ensure code reusability, HET can benefit from Turing Complete structure\cite{turing1936computable}. The structure of HET is an enhanced preorder traversal finite state machine --- a  tree composed of various node types, supporting arbitrary jumps between  nodes and dynamic structural modifications. Node types include: atomic operations (e.g., moving to a position, capturing an image, invoking basic tool algorithms like Grounding-DINO), composite operations (e.g., grasping an object), conditional branches (IF operations), jump operations (e.g., jump to Node A), exit operations (terminating the traversal loop).

By logically combining these functions, the module effectively expresses complex task-handling logic, achievable through proper prompt engineering  for the VLM. This approach improves function reuse rates, reduces the  number of calls to large models, and enhances task execution efficiency.  
%Experiments show that in industrial assembly tasks, the module reuse  rate of the task tree reaches 62\%, and new task development efficiency improves by 40\% (see Chapter 4 for results).

\subsection{Multi-level Anomaly Handling(MAH)}
\label{3.3}

In robotic manipulation tasks, anomaly handling refers to the timely detection and correction of unexpected events or disruptions during execution, including deviations from the planned actions, accidental object drops, environmental disturbances, and other unanticipated scenarios. Formally, given an initial execution tree $V_i=\{v_j\}_{j=1}^{M_i}$, the system computes low-dimensional representation of current observation $s_t=f(O(t))$, where $f:\mathbb{R}^{H\times W\times 4}\to \mathbb{R}^{P_t\times3}$, which are the spatial positions of the keypoints proposed by the VLM, and computes the expected current scene state $u_{t}$ from VLM-defined functions. Then the system applies an anomaly detection function
% \[
% \delta_t = \mathcal{E}\bigl(O_t,\,\sigma^0_t\bigr),
% \]
\[
\delta_t = \mathcal{E}\bigl(s_t,\,u_{t}\bigr),
\]
where \(\delta_t > \tau_i\) indicates a detected anomaly. Then the correction module then update the call sequence, ensuring task continuity and success. Anomaly handling is essential for improving a robot's robustness and reliability in dynamic, unstructured environments.

There exist myriad approaches to anomaly detection and correction, as this is a broad and extensively research area. Different methods vary in their anomaly-handling capabilities, the types and scope of anomalies they can address, and the resources required for their execution. Striking a balance between success rate and resource cost in anomaly handling is a critical challenge.

%Through extensive statistical analysis of execution data in our desktop experimental platform, we categorize anomalies into three difficulty levels: predictable anomalies (46\%), which can be rapidly detected and corrected using predefined rules; rearrange recoverable anomalies (37\%), which can be resolved through localized adjustments to the current task sequence without full replanning; and complex anomalies (17\%), whose degree of anomaly complexity or workspace changes necessitate global replanning via the VLM.

Through extensive statistical analysis of execution data in our desktop experimental platform, we categorize anomalies into three difficulty levels: predictable anomalies, which can be rapidly detected and corrected using predefined rules; rearrange recoverable anomalies, which can be resolved through localized adjustments to the current task sequence without full replanning; and complex anomalies, whose degree of anomaly complexity or workspace changes necessitate global replanning via the VLM.

Based on this insight, we propose a multi-level anomaly handling framework consisting of three components. First, \textbf{predefined monitors} are generated by the VLM at the start of the pipeline to address Predictable Anomalies. These handlers continuously monitor key variables in the scene metadata and detect any predefined anomalies. Second, a lightweight, fine-tuned\textbf{ local anomaly expert} implemented as a small language model handles rearrange recoverable anomalies. When an anomaly escapes the predefined handlers, this expert attempts to adjust the call sequence based on the scene metadata. Third, cloud-based\textbf{ VLM replanning} tackles complex anomalies that neither predefined handlers nor local adjustments can resolve. The entire framework is implemented as hierarchical executable code generated by the VLM, which runs in a separate thread from the main control logic and invokes the underlying interfaces. We discuss these three components in more detail in Appendix\ref{More details about components of Multi-level Anomaly Handling (MAH)}.

% , the correction module consists of 3 level of anomaly handing techniques: predefined monitors, local anomaly expert and VLM replaning, whose anomaly handing logic are all generated by VLM.

% \paragraph{Metadata.}
% Metadata refers to scene information that is collected in real time and structured via a predefined visual pipeline; detailed methodology and examples are provided in the Appendix.

In implementation, the predefined monitors are realized as executable code generated directly by the VLM, while both the local anomaly expert and the VLM replanning function are exposed as interfaces within the skill library. Additionally, the VLM produces a scheduler to orchestrate all three modules into a complete multi-level anomaly handling framework, which executes in a separate thread at runtime.

The multi-level anomaly handling framework stratifies anomalies by difficulty, enabling rapid recovery of simple faults and cautious replanning for complex failures, mirroring human decision patterns and striking an effective balance between success rate and response latency. As an adaptive paradigm, it dynamically routes each anomaly to the most suitable processing tier based on its predictability and complexity, achieving both low-latency local recovery and high-robustness global corrections. Its hierarchical design is inspired by the human nervous system's feedback loops (spinal cord $\to$ cerebellum $\to$ cortex), ensuring overall performance while maintaining efficiency and reliability.

\subsection{Hierarchical Incremental Memory Management (HIMM)}
\label{3.4}
% In current AI agent-related work, a major factor limiting operational speed is the repeated calls to VLMs. Additionally, practical experience in building AI agents has shown that many generated functions during task execution are redundantly regenerated. Experiments indicate that if functions can be reused, the output of large models can be reduced \textcolor{red}{(see ablation)}. To fully leverage each output of large models, minimize the number of calls, and endow the model with embodied knowledge after each invocation, we propose the HIMM module.  

% As is shown in \textcolor{red}{[Figure X]}. HIMM manages a hierarchical memory module (HMM) and an Incremental Skill Pool (ISP). The HMM consists of short-term memory, medium-term memory, and lifelong memory. It continuously records function invocations and anomaly handling logs in short-term memory, and periodically summarized by a locally deployed large model  into medium-term memory (frequently called functions) and lifelong memory (prompt-level optimizations for embodied intelligence based on task execution experience).  

% The incremental skill pool contains atomic skills and composite skills, with their function headers and descriptions appended to prompts during each invocation. After multiple calls, frequently used functions from medium-term memory are added to the skill pool, improving code reuse efficiency.  

% ------ DUPLICATED???

The operational efficiency of current AI agent systems is constrained by two key factors: First, frequent calls to Vision-Language Models (VLMs) create a significant performance bottleneck. In typical task execution, approximately 43\% of the latency stems from serialized VLM queries (see ablation). Second, task trajectory analysis reveals that up to 48\% of large model outputs consist of functionally equivalent but textually varied duplicate function generations (see ablation). Thus, an effective function reuse mechanism can reduce large model invocation frequency, leading us to propose the Hierarchical Incremental Memory Management (HIMM) module.  

 % As is shown in \textcolor{red}{[Figure X]}, the core architecture of HIMM consists of two synergistic subsystems: The hierarchical memory module (HMM) employs a three-tier structure for knowledge accumulation—short-term memory acts as a rolling-window cache (capacity $\leqslant$ 10 entries), recording raw task trajectories and exception logs in real time; medium-term memory uses a lightweight local model [energy consumption only 5\% of VLM] to cluster high-frequency function templates (invoked $\geqslant$ 3 times/hour) via LRU summarization every 10 minutes; lifelong memory stores human-verified embodied prompt optimizations (e.g., "keep the wrist horizontal during robotic arm grasping").  
% As is shown in \textcolor{red}{[Figure X]}. 
HIMM manages a Hierarchical Memory Module (HMM) and an Incremental Skill Pool (ISP). The HMM consists of short-term memory, medium-term memory, and lifelong memory. It continuously records function invocations and anomaly handling logs in short-term memory, and periodically summarized by a locally deployed large model  into medium-term memory (frequently called functions) and lifelong memory (prompt-level optimizations for embodied intelligence based on task execution experience).  

\textbf{Hierarchical Memory Module (HMM)}: HMM employs a three-tier structure for knowledge accumulation --- short-term memory acts as a rolling-window cache (capacity $\leqslant$ 10 entries), recording raw task trajectories and exception logs in real time; medium-term memory uses a lightweight local model to cluster high-frequency function templates (invoked $\geqslant$ 3 times/hour) via LLM summarization every 10 minutes; lifelong memory stores human-verified embodied prompt optimizations (e.g., "keep the wrist horizontal during robotic arm grasping").  

\textbf{Incremental Skill Pool (ISP)}: The ISP implements a two-tier skill system: atomic skills (e.g., \texttt{move\_to(x,y,z)}) as indivisible action primitives, and composite skills (e.g., \texttt{package\_sorting())} described via directed acyclic graphs. Its knowledge accumulation follows strict quality control: when medium-term memory detects a function invoked over five times, semantic normalization eliminates superficial variations, and only those with $\geqslant$ 90\% test coverage are added. 

This design offers two key advantages: First, a dynamic prompt mechanism automatically injects skill context, such as appending available skill descriptions and relevant memory scenarios to each VLM call. Second, an exception-driven update strategy logs failures in short-term memory while triggering revalidation of related skills.  

%Tests on robots \textcolor{red}{(See Sec xxx)} show HIMM increases skill reuse from 17\% to 63\%, reduces VLM calls by 29\%, and improves task success rates by 11\%, while lifelong memory reduces the time to get used to new tasks by 40\%. This dual-channel optimization of memory and skills provides a verifiable solution to overcoming AI agents' "regeneration trap."

Tests on robots (See Sec. \ref{4.3}) show HIM reduces generated code size by 41\%, reduces VLM calls by 27\%. This dual-channel optimization of memory and skills provides a verifiable solution to overcoming AI agents' "regeneration trap."

\section{Experiments}
\label{headings}
\label{Experiments}

In this section, we perform experiments to investigate four main research questions: (1) What are the performance and efficiency metrics of our system on open‐vocabulary manipulation tasks across various real‐world scenarios? (Sec. \ref{Real-World and  Open-Vocabulary Manipulation}); (2) Can our system achieve the anomaly detection and recovery capabilities and the cross‐task long‐term memory performance claimed in Section 3? (Sec. \ref{4.2}); (3) What is the quantitative contribution of each functional module to the overall system performance? (Sec. \ref{4.3}); and (4) How significantly do individual modules contribute to task failures, and what are their main error sources and failure patterns? (Sec. \ref{4.4}).

To evaluate our method’s real‐world performance, we deployed a desktop experimental setup in an actual environment: an Intel RealSense D435i RGB‑D camera was mounted above the desktop to capture scene observations, and a UR5e 6‑DoF robotic arm equipped with a 1‑DoF gripper performed the manipulation tasks. Our experiments are conducted on an Intel(R) Core(TM) i7-14700KF CPU and an NVIDIA RTX A6000 GPU.

\subsection{Real-World and  Open-Vocabulary Manipulation}
\label{Real-World and  Open-Vocabulary Manipulation}

\begin{table}[ht]
	\centering
	\caption{Quantitative evaluation of manipulation performance across 10 task instances per method. Success rates (successful trials/total trials) and average execution times are reported. FrankenBot achieves significantly higher success rates (73\% overall), demonstrating robust task adaptation capabilities.}
	\label{tab:quant-15}
	\begin{tabular}{p{4.5cm}cccccc}
		\toprule
		\textbf{Task} & \multicolumn{2}{c}{\textbf{VoxPoser}} & \multicolumn{2}{c}{\textbf{ReKep}} & \multicolumn{2}{c}{\textbf{FrankenBot (Ours)}} \\
		\cmidrule(lr){2-3} \cmidrule(lr){4-5} \cmidrule(lr){6-7}
		& Success & Time & Success & Time & Success & Time \\
		\midrule
		Block Picking              & 7/10 & 12.4s  & 7/10 & 20.8s   & \textbf{9/10} & \textbf{7.2s} \\
		Block Picking \& Stacking  & 5/10 & 18.7s  & 6/10 & 35.2s  & \textbf{9/10} & \textbf{10.5s} \\
		Drawer Closing             & 8/10 & 8.5s   & 10/10 & 17.1s  & \textbf{10/10} & \textbf{5.3s} \\
		Tea Pouring                & 1/10 & 25.3s  & 3/10 & 20.6s  & \textbf{5/10} & \textbf{16.8s} \\
		Pen Reorienting           & 4/10 & 10.2s  & 5/10 & 20.9s  & \textbf{8/10} & \textbf{6.7s} \\
		Trash Sorting              & 2/10 & 26.8s  & 4/10 & 59.5s  & \textbf{7/10} & \textbf{21.4s} \\
		Toy Storing             & 6/10 & 15.3s  & 6/10 & 25.3s  & \textbf{7/10} & \textbf{8.9s} \\
		Button Pressing           & 8/10 & 7.2s   & 8/10 & 13.4s  & \textbf{10/10} & \textbf{5.2s} \\
		Shoes Packing             & 5/10 & 22.1s  & 5/10 & 24.7s  & \textbf{6/10} & \textbf{20.3s} \\
		USB Plugging              & 0/10 & N/A    & 1/10 & 62.7s  & \textbf{2/10} & \textbf{44.6s} \\
		\midrule
		\textbf{Total}            & \multicolumn{2}{c}{46\%} & \multicolumn{2}{c}{55\%} & \multicolumn{2}{c}{\textbf{73\%}} \\
		\bottomrule
	\end{tabular}
\end{table}

We designed and selected ten manipulation tasks derived from everyday real-world scenarios, covering  simple move tasks to more challenging tasks that include multiple steps with high failure probability or require cross‑task contextual memory. For each task, we ran ten independent trials, randomly initializing the poses of all task-relevant objects in the scene for each trial. Performance was evaluated using two metrics: task success rate and completion time. We chose VoxPoser and ReKep as our baselines for comparison.

Table \ref{tab:quant-15} details the quantitative results. We observe that our method demonstrates robust zero‑shot generalization, delivering particularly impressive performance on long‑horizon tasks, with significant gains over the baselines on both success rate and completion time. Unlike fixed anomaly‑handling schemes, where the task is partitioned into rigid stages and errors simply trigger a rollback to the previous stage, our multi‑level anomaly‑handling mechanism adaptively allocates appropriate resources to different failure modes, offering a far more flexible and powerful paradigm. This design equips our system with both exceptional recovery capabilities and, correspondingly, high execution efficiency, in full agreement with the analysis in Section 3. Moreover, we find that our approach exhibits cross‑task foresight, leveraging information accrued during prior task executions to guide current operations. This benefit originates from our multi‑level memory architecture, which actively retrieves relevant memories at each execute to support decision‑making. We will discuss the importance of these components in our framework's performance in the ablation study of Sec \ref{4.3}.

\subsection{Anomaly Handling and Cross‑Task Memory}
\label{4.2}
	\par Prior to formal evaluation, we conducted foundational experiments to establish the natural occurrence rates of different anomaly classes in real-world manipulation scenarios. Five representative tasks were selected spanning household activities (block stacking, drawer closing, tea pouring, object sorting, and pen reorienting), with each task executed across 10 trials under varying environmental conditions.
%	\begin{table}[!ht]
%		\centering
%		\caption{Natural vs. Calibrated Anomaly Distribution Across Tasks}
%		\label{tab:anomaly_distribution}
%		\small
%		\begin{tabular}{lcccccc}
%			\toprule
%			\textbf{Task} & \textbf{Trials} & \multicolumn{3}{c}{\textbf{Natural Occurrence}} & \multicolumn{2}{c}{\textbf{Calibrated Test}} \\
%			\cmidrule(lr){3-5} \cmidrule(lr){6-7}
%			&  & Predictable & Recoverable & Complex & Level & Ratio \\
%			\midrule
%			Block Stacking & 10 & 8 (80\%) & 1 (10\%) & 1 (10\%) & P & 5 \\
%			Drawer Closing & 10 & 5 (50\%) & 3 (30\%) & 2 (20\%) & R & 3 \\
%			Tea Pouring & 10 & 7 (70\%) & 2 (20\%) & 1 (10\%) & P & 5 \\
%			Button Pressing & 10 & 6 (60\%) & 3 (30\%) & 1 (10\%) & R & 3 \\
%			Pen Reorienting & 10 & 9 (90\%) & 1 (10\%) & 0 (0\%) & C & 2 \\
%			\midrule
%			\textbf{Total} & 50 & 35 (70\%) & 10 (20\%) & 5 (10\%) &  & 5:3:2 \\
%			\bottomrule
%		\end{tabular}
%	\end{table}
	
	\begin{table}[!ht]
		\centering
		\caption{Natural Anomaly Occurrence Statistics Across Tasks}
		\label{tab:natural_anomaly_distribution}
		\small
		\begin{tabular}{lcccccc}
			\toprule
			\textbf{Task (Difficulty)} & \textbf{Trials} & \textbf{Total} & \multicolumn{3}{c}{\textbf{Anomaly Type Count}} & \textbf{Avg. per} \\
			& & \textbf{Anomalies} & \textbf{Predictable} & \textbf{Recoverable} & \textbf{Complex} & \textbf{Trial} \\
%			\cmidrule(lr){4-6}
			\midrule
			Block Stacking (Easy) & 10 & 12 & 10 (83.3\%) & 2 (16.7\%) & 0 (0\%) & 1.2 \\
			Drawer Closing (Easy) & 10 & 17 & 13 (76.5\%) & 3 (17.6\%) & 1 (5.9\%) & 1.7 \\
			Tea Pouring (Hard) & 10 & 47 & 28 (59.6\%) & 12 (25.5\%) & 7 (14.9\%) & 4.7 \\
			Button Pressing (Easy) & 10 & 9 & 8 (88.9\%) & 1 (11.1\%) & 0 (0\%) & 0.9 \\
			Pen Reorienting (Medium) & 10 & 25 & 17 (68.0\%) & 6 (24.0\%) & 2 (8.0\%) & 2.5 \\
			\midrule
			\textbf{Total} & 50 & 110 & 76 (69.1\%) & 24 (21.8\%) & 10 (9.1\%) & 2.2 \\
			\bottomrule
		\end{tabular}
	\end{table}
	
	\par The initial analysis revealed a strongly skewed natural distribution - approximately 70\% predictable anomalies (largely object pose deviations and computer vision inaccuracy), 20\% recoverable anomalies, and only 10\% complex anomalies (system-level failures or inappropriate planning), as is demonstrated in Table \ref{tab:natural_anomaly_distribution}. While this reflects real-world frequencies, it presents challenges for comprehensive system evaluation as complex cases become statistically insignificant. We therefore designed the test distribution to 5:3:2 through controlled anomaly injection, with 10 trials per anomaly type. As is shown in Table \ref{tab:anomaly_performance}, performance was evaluated using recovery success rate and time penalty. The baselines we chose are naive rollback, full remote VLM replan and local LLM replan.

    % while maintaining ecological validity via three safeguards: (a) all injected anomalies were derived from real failure cases in our robot deployment logs, (b) anomaly difficulty was validated by three independent robotics experts (Cohen's $\kappa=0.79$), and (c) physical feasibility was confirmed through Gazebo simulations.

\begin{table}[tb]
	\centering
%	\caption{Performance comparison of anomaly handling strategies: (1) naive rollback, (2) full VLM replanning, (3) local LLM correction, and (4) our MAH framework. The 5:3:2 test distribution reflects real-world anomaly ratios observed in household manipulation tasks. MAH reduces latency by 61\% vs. full replanning (7.2s vs. 18.5s avg.) while achieving 97\% overall success rate.
%		\\		SR: Success Rate, $\Delta t$: Time penalty (recovery time - nominal time)}
	\caption{Performance comparison of anomaly handling strategies. SR: Success Rate, $\Delta t$: Time penalty (recovery time - nominal time)}
	\label{tab:anomaly_performance}
%	\scriptsize
	\setlength{\tabcolsep}{2pt}
	\begin{tabular}{llcccccccc}
		\toprule
		\textbf{Class} & \textbf{Anomaly Example} & \multicolumn{2}{c}{\textbf{Naive Rollback}} & \multicolumn{2}{c}{\textbf{Full Replan}} & \multicolumn{2}{c}{\textbf{Local LLM}} & \multicolumn{2}{c}{\textbf{MAH (Ours)}} \\
		\cmidrule(lr){3-4} \cmidrule(lr){5-6} \cmidrule(lr){7-8} \cmidrule(lr){9-10}
		& & SR & $\Delta t$ & SR & $\Delta t$ & SR & $\Delta t$ & SR & $\Delta t$ \\
		\midrule
		
		\multirow{5}{*}{\rotatebox[origin=c]{0}{\textbf{Predictable}}}
		
		& Object displacement & 45\% & 0.7s & 100\% & 11.2s & 95\% & 2.8s & 100\% & 0.2s \\
		& Sensor noise & 50\% & 0.8s & 100\% & 10.8s & 90\% & 3.1s & 100\% & 0.3s \\
		& Minor pose deviation & 40\% & 0.9s & 95\% & 12.5s & 85\% & 3.5s & 95\% & 0.4s \\
		& Expected collision & 55\% & 0.6s & 100\% & 9.7s & 92\% & 2.6s & 100\% & 0.3s \\
		& Temporary occlusion & 35\% & 1.0s & 90\% & 13.8s & 80\% & 4.0s & 90\% & 0.5s \\
		\cmidrule(l){3-10}
		& \textbf{Average} & 45\% & 0.8s & 97\% & 11.6s & 88\% & 3.2s & \textbf{97\%} & \textbf{0.3s} \\
		
		\midrule
		
		\multirow{3}{*}{\rotatebox[origin=c]{0}{\textbf{Recoverable}}}
		
		& Gripper slip & 25\% & 4.7s & 95\% & 14.5s & 80\% & 3.7s & 95\% & 2.6s \\
		& Partial object drop & 30\% & 4.3s & 90\% & 15.1s & 75\% & 4.2s & 90\% & 2.9s \\
		& Mid-task obstruction & 20\% & 5.0s & 85\% & 16.8s & 70\% & 5.0s & 85\% & 3.5s \\
		\cmidrule(l){3-10}
		& \textbf{Average} & 25\% & 4.7s & 90\% & 15.5s & 75\% & 4.3s & \textbf{90\%} & \textbf{3.0s} \\
		
		\midrule
		\multirow{2}{*}{\rotatebox[origin=c]{0}{\textbf{Complex}}}
		& Topology change & 5\% & N/A & 65\% & 29.8s & 60\% & 10.2s & 80\% & 19.5s \\
		& Cross-task conflict & 0\% & N/A & 55\% & 32.1s & 50\% & 12.0s & 75\% & 22.3s \\
		\cmidrule(l){3-10}
		& \textbf{Average} & 3\% & N/A & 60\% & 31.0s & 55\% & 11.1s & \textbf{78\%} & \textbf{20.9s} \\
		
		\bottomrule
	\end{tabular}
	\vspace{0.2cm}
\end{table}
	% \par We designed a stress-test protocol injecting three anomaly classes (predictable, recoverable, and complex) during standardized manipulation tasks (cup stacking and drawer closing), with 10 trials per anomaly type. 

\subsection{Ablation Study}
\label{4.3}

We conducted ablation studies to evaluate the contribution of each key component to overall system performance. These experiments employed the same task set and evaluation metrics (success rate and average completion time) as defined in Sec. \ref{Real-World and  Open-Vocabulary Manipulation} and \ref{4.2}. Given that the functionality of MAH was already validated in Sec \ref{4.2}, we focused on two variants: (1) No hierarchical memory module: The VLM receives only the current instruction and scene observation, without access to historical experience prompts. (2) No execution tree, where the agent can only execute the code from serial function series. (3) No incremental skill library: The library remains static after initialization, prohibiting dynamic expansion or functional reuse.

To isolate the impact of each component, we conducted 10 consecutive block-stacking trials for 3 runs under two ablated conditions: (1) Memory-less execution: The system processed only real-time visual observations and immediate instructions, with no access to hierarchical memory. This tested the baseline capability without experiential learning. (2) Static skill library: The basic skill library remained frozen during trials, preventing dynamic updates from task progression.

%\begin{table}[ht]
%	\centering
%	\caption{Ablation Study of System Components (Lower is Better). Each trial’s outcome was evaluated using success rate (stacking completion) and efficiency (average VLM calls, code length and execution time). The consecutive trial design exposed compounding effects of memory/skill limitations, as errors accumulated without corrective adaptation.}
%	\label{tab:ablation}
%	\small
%	\begin{tabular}{lcccc}
%		\toprule
%		\textbf{Configuration} & \textbf{Success Rate} & \textbf{VLM Calls} & \textbf{Code Size (LOC)} & \textbf{Exec. Time (s)} \\
%		\midrule
%		Full Model & 70\% & 1.1 & 58 & 12.3 \\
%		\midrule
%		w/o HIM & 30\% & 5.7 & 82 & 58.6 \\
%		w/o Execution Tree & 60\% & 1.3 & 112 & 17.4 \\
%		w/o ISP & 80\%  & 1.5 & 67 & 13.2 \\
%		\bottomrule
%	\end{tabular}
%\end{table}

\begin{table}[ht]
	\centering
	\caption{Ablation study of system components (lower is better). The evaluation metrics include success rate, computational cost (VLM calls and code size), and detailed time breakdown. Time is split into \textbf{VLM/LLM Generation} (prompt processing and planning) and \textbf{Execution} (physical operation time). Results are reported over 3 runs different seeds. }
	\label{tab:ablation1}
	\small
	\begin{tabular}{lccccc}
		\toprule
		\textbf{Configuration} & \textbf{Success} & \textbf{VLM} & \textbf{Code Size} & \multicolumn{2}{c}{\textbf{Time (s)}} \\
		
		\cmidrule(lr){5-6}
		& \textbf{Rate} & \textbf{Calls} & \textbf{(LOC)} & \textit{Generation} & \textit{Execution} \\
		\midrule
		Full Model            & $70\%_{\scriptstyle\pm10\%}$  & 1.1  & 58  & 9.2 & 12.1  \\
		\midrule
		w/o HMM              & $26.7\%_{\scriptstyle\pm4.71\%}$  & 5.7  & 82  &  60.2 & 68.4 \\
		w/o Execution Tree   & $63.3\%_{\scriptstyle\pm4.71\%}$  & 1.3  & 112 & 12.1 & 13.3 \\
		w/o ISP              & $73.3\%_{\scriptstyle\pm12.47\%}$  & 1.5  & 67  & 13.8 & 15.4  \\
		\bottomrule
	\end{tabular}
\end{table}

% $70\%_{\scriptstyle\pm10\%}$

%Further details on how alternative design options impact overall system performance are provided in the appendix.

%Table 2 summarizes the results. First, incorporating the multistage memory module yields clear cross-task generalization benefits: prior successes and failures reliably improve system performance, and tasks demanding cross-task context see particularly large gains. Second, without the multistage exception handler, the system must trade off success rate against execution efficiency, which aligns with our analysis in Section 3. Finally, enabling skill library updates produces a noticeable improvement in overall success rate; while the improvement is modest, it is nonetheless statistically significant.

\subsection{System Error Breakdown}
\label{4.4}

\begin{figure}[!ht]
    \centering
    \includegraphics[width=0.5\linewidth]{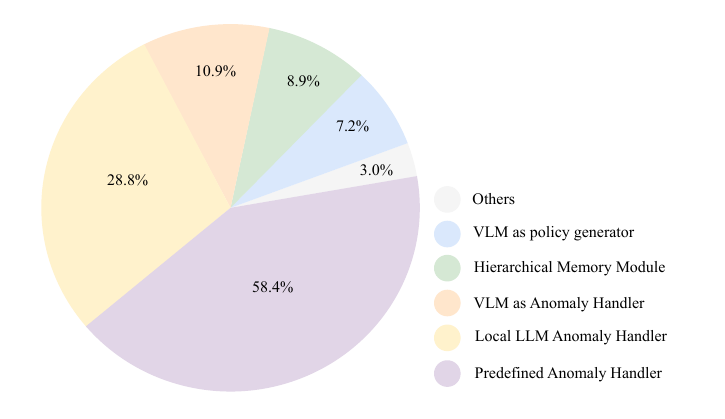}
    \caption{Pie chart of error type distribution}
    \label{fig:bingtu}
\end{figure}

By virtue of the interpretability and modular design of our framework, we can pinpoint and quantify component‑wise error rates, as shown in Fig. \ref{fig:bingtu}. First, although the error rate of the VLM policy generation module is generally acceptable, its performance still depends on the VLM’s inference capabilities. We found that our method relies on more powerful VLMs such as GPT‑4.1. The Hierarchical Memory Module, by contrast, is highly stable, as it only needs to maintain simple task and scene descriptions.

% Within the Multi-level Anomaly Handling module, predefined, condition‑triggered detectors activate only when specific conditions are met, thereby delivering reliable performance with a low failure rate. 
While predefined condition-triggered detectors work well for simple cases, their fixed logic struggles with complex real-world situations where anomalies often need more flexible responses. Attempting to use a local lightweight local LLM for anomaly handling, however, leads to noticeably higher error rates, owing to its limited scale and commonsense reasoning capacity. In comparison, VLM‑based error correction leveraging a cloud‑scale model to replan and reexecute demonstrates greater robustness, though at the cost of increased execution time.

Overall, apart from these key modules, the remaining components have a minimal impact on system stability.

\section{Conclusion}
\label{Conclusion}

In this work, we present a VLM-driven, brain-like framework for robotic manipulation that realizes a functionally comprehensive robotic brain while maintaining high operational efficiency. Our method offers several key advantages: first, the framework is plug-and-play --- deployment requires only implementation of the defined interfaces, with no additional training; second, it constitutes a functionally complete robotic brain by integrating core features such as task planning, anomaly detection and correction, and long-term memory; and finally, it operates with high efficiency, performing nearly every task with just a single VLM call. Moreover, the framework’s modules are highly decoupled, enabling rapid integration of advanced techniques—such as VLM applications in robotics, code generation, and RAG (retrieval-augmented generation) techniques—and, if finer-grained scene perception and motion control are provided at the low-level interfaces, the framework will immediately realize corresponding performance gains. While advantageous, our method still has several limitations. In Appendix \ref{limi}, we discuss the limitations of FrankenBot and outline several directions for future work.

% \section*{References}

% {
% \small

% [1] Huang, W., Wang, C., Li, Y., Zhang, R., \& Fei-Fei, L. (2024). Rekep: Spatio-temporal reasoning of relational keypoint constraints for robotic manipulation. arXiv preprint arXiv:2409.01652.

% [2] Huang, W., Wang, C., Zhang, R., Li, Y., Wu, J., \& Fei-Fei, L. (2023). Voxposer: Composable 3d value maps for robotic manipulation with language models. arXiv preprint arXiv:2307.05973.

% [3] Pan, M., Zhang, J., Wu, T., Zhao, Y., Gao, W., \& Dong, H. (2025). OmniManip: Towards General Robotic Manipulation via Object-Centric Interaction Primitives as Spatial Constraints. arXiv preprint arXiv:2501.03841.

% [4] Huang, H., Lin, F., Hu, Y., Wang, S., \& Gao, Y. (2024, October). Copa: General robotic manipulation through spatial constraints of parts with foundation models. In 2024 IEEE/RSJ International Conference on Intelligent Robots and Systems (IROS) (pp. 9488-9495). IEEE.

% }

% \cite{huang2023voxposer}

\bibliographystyle{unsrt}
\bibliography{reference}

%%%%%%%%%%%%%%%%%%%%%%%%%%%%%%%%%%%%%%%%%%%%%%%%%%%%%%%%%%%%

\newpage

\appendix

\section{Technical Appendices and Supplementary Material}

\subsection{Broader Impacts}
\label{Broader}

FrankenBot can be used in manufacturing to improve the accuracy and flexibility of automated production and reduce repetitive labor. FrankenBot enhances robot’s capabilities, but it could be repurposed for negative applications such as advanced autonomous weapon development or pose safety risks if VLM inferences are flawed. To mitigate these hazards, we recommend incorporating human-in-the-loop verification for high-risk tasks and establishing deployment monitoring and feedback mechanisms to ensure responsible use.

\subsection{Extended Discussion of Limitations and Future Works}
\label{limi}

While advantageous, our method still has several limitations. The multi-threaded executable code generation depends on the VLM’s strong reasoning capabilities, and our experiments show that it only maintains ideal performance when using newer versions of the VLM. Fine-grained scene understanding and low-level motion control fall outside the scope of this work and instead rely entirely on low-level interface implementations, which may become performance bottlenecks in real-world applications. Our multi-level anomaly handling system is based on simplistic assumptions about task setups—while it performs well in relatively structured scenarios (e.g., desktop manipulation), its effectiveness may degrade significantly in more complex environments. Finally, because we propose only the framework, the current memory management mechanism is rather rudimentary, and long-duration, consecutive task executions may suffer from information redundancy or forgetting.

The rapid advances in related fields have opened up several exciting directions for future work. One active area is to collect large-scale embodied task–related data and perform domain-adaptive fine-tuning of pretrained large VLMs, in order to endow them with spatial reasoning capability, domain priors, and semantic grounding capability better suited to robotic manipulation. Retrieval-augmented generation (RAG) techniques likewise constitute a vast and dynamic research domain, the latest advances in this field warrant exploration for applications in robot long-term memory. Finally, in the Vision-Language-Action (VLA) arena, gathering larger embodied datasets and designing more advanced architectures to achieve more robust, generalizable low-level motion control—and rapid handling of execution anomalies to cope with changing scenarios—also represents a highly promising research direction.

% \subsection{Real-World Environment Setup}

\subsection{Hardware Setup}

\begin{figure}[!ht]
    \centering
    \includegraphics[width=\linewidth]{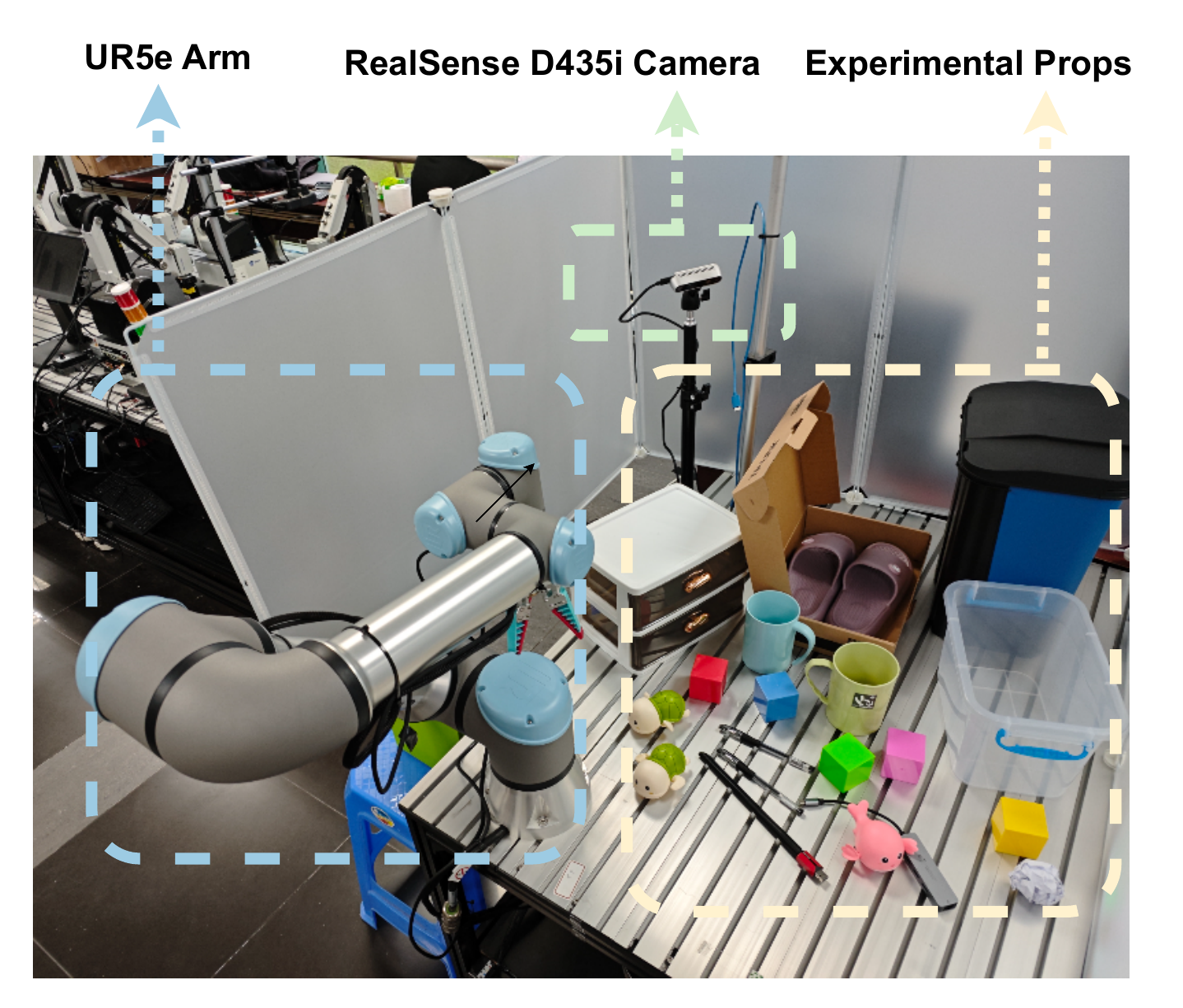}
    \caption{Execution Snapshot for Experimental Tasks}
    \label{fig:family}
\end{figure}

As shown in Fig. \ref{fig:family}, we deploy a UR5e 6-DoF arm on our desktop platform, outfitted with a stepper-motor-driven 1-DoF gripper for object grasping. Robot control from the host PC is implemented via URScript in conjunction with the RTDE interface, supporting a theoretical maximum communication rate of 500 Hz. Scene perception relies solely on the depth and color streams of a single Intel RealSense D435i RGB-D camera, demonstrating the low sensor-hardware requirements of our approach in practical deployment.

\subsection{Connections and Distinctions between FrankenBot and existing works
}
\label{arw}

\paragraph{Brain-like structures in robotic manipulation tasks. }
Our method differs fundamentally from existing approaches: whereas prior work relies on data collection and fine‑tuning to transform a pre‑trained VLM into a VLA capable of end‑to‑end action sequence inference, our approach retains the VLM at its core without any additional training. Instead, we ground the VLM's reasoning in embodied tasks through a suite of highly structured modules such as a skill library, multi‑level anomaly handling, and multi‑level memory to construct a fully functional and efficiently robotic brain.

\paragraph{Vision-Language Models for Robotics.} 
The key difference between our method and VLMs for robotics is that, while existing methods typically address only a single "brain" module, such as planning, policy generation, or fault recovery, our framework realizes a more complete, brain‑inspired VLM‑driven manipulation system that integrates all major capabilities while remaining resource‑efficient.

% \subsubsection{Tasks}

% \subsubsection{Baseline Methods}

% \subsection{Simulated Environment Setup}

% \subsubsection{Environment Setup}

% \subsubsection{Tasks}

% \subsubsection{Baseline Methods}

% \subsection{Details of Constraints}

% \subsection{Comparison of methods for selecting spatial representation extraction tools}

% \subsection{VLM and prompts}

\subsection{Comparison of VLMs}
\label{apvlm}

\begin{table}[ht]
  \centering
  \caption{Exploratory study of VLM choice effects on system performance}
  \label{tabVLM}
  \begin{tabular}{lcc}
    \toprule
    \textbf{Model}               & \textbf{Inference Validity Rate (\%)} & \textbf{Total Success Rate (\%)} \\
    \midrule
    o3                           & 87                                & 73                              \\
    GPT-4.1                      & 84                                & 71                              \\
    GPT-4o mini                  & 81                                & 65                              \\
    gpt-4-vision-preview         & 80                                & 63                              \\
    \bottomrule
  \end{tabular}
\end{table}

To investigate how model choice affects overall system performance, we designed an exploratory experiment using all tasks from the comparative study in Sec. \ref{Real-World and  Open-Vocabulary Manipulation}. Experimental results are presented in Tab. \ref{tabVLM}. We observe that lower-tier models achieve overall acceptable performance, demonstrating our method’s robustness to model variation, yet more advanced, newer models deliver clearly superior results across both evaluation metrics. This finding conveys two insights. First, our approach will continue to benefit as VLM reasoning capabilities improve in the future. Second, there is a promising research direction in designing system architectures that more effectively structure and guide the VLM’s reasoning process so as to lessen dependence on its peak performance. For these reasons, we ultimately selected GPT-4.1 as the VLM for our method: it delivers nearly the best reasoning performance while costing only about twenty percent of o3.

% \subsubsection{prompts}

% \subsection{Details of ToolKit}

% \subsubsection{Samples of the toolkit we provide}

% \subsubsection{Evaluation and replacement of tools}

% \subsection{Details of  Fine-grained spatial representation extraction process}

% \subsubsection{Details of extraction process}

% \subsubsection{Comparison of different query forms}

% \subsection{More Visualization}

% \subsection{Encapsulated as an Agent}

\subsection{More details about components of Multi-level Anomaly Handling (MAH)}
\label{More details about components of Multi-level Anomaly Handling (MAH)}

\textbf{Predefined Monitors}: Inspired by prior work that uses constraint code to reason about tracked elements in the scene, our predefined monitors follow a similar structural approach. During the planning phase, we invoke the VLM to infer a set of the most common and easily rule-based anomalies (e.g., "object unexpectedly moved"). For each anomaly pattern, the VLM automatically generates two Python routines: a monitor that continuously observes key variables in the scene metadata (such as object position), and a handler that, upon anomaly detection, executes a predefined rollback or compensation strategy  (for example, re-localizing the object via vision and retrying the grasp). At runtime, the monitor thread runs in parallel with the control thread, enabling low latency, localized handling of predictable anomalies.

\textbf{Local Anomaly Expert}: When an anomaly is not caught by the predefined monitors, the system switches to a local anomaly expert, which is a fine-tuned small language model. The expert analyzes scene metadata and the current task instruction to infer the anomaly cause (for example, "the bottle count changed from one to two"). It then determines whether the anomaly falls within its "sequence-reorganization" recovery scope: if so, it inserts or adjusts steps in the skill-call sequence (for example, transforming "locate object $\to$ move $\to$ grasp" into "locate object $\to$ move $\to$ grasp $\to$ locate object $\to$ move $\to$ grasp"); if not, the anomaly is escalated to the VLM replanning module. This mechanism bridges the gap between predefined rule-based recovery and VLM full replanning, achieving a balance between response speed and anomaly-handling capability. 

\textbf{VLM Replanning}: For complex anomalies that exceed the local expert's capabilities, we encapsulate the current scene state and task instruction into a prompt for the VLM, which then generates a corrected executable code sequence. Although this procedure introduces higher latency, it ensures system robustness and correctness in the most extreme or highly uncertain scenarios.

\end{document}